\documentclass{article}

\usepackage[preprint]{neurips_2025}

\usepackage[utf8]{inputenc} % allow utf-8 input
\usepackage[T1]{fontenc}    % use 8-bit T1 fonts
\usepackage{hyperref}       % hyperlinks
\usepackage{url}            % simple URL typesetting
\usepackage{booktabs}       % professional-quality tables
\usepackage{amsfonts}       % blackboard math symbols
\usepackage{nicefrac}       % compact symbols for 1/2, etc.
\usepackage{microtype}      % microtypography
\usepackage{xcolor}         % colors
\usepackage{amsmath} 
\usepackage{amssymb}
\usepackage{multirow}
\usepackage{graphicx}
\usepackage{CJKutf8}

\title{Identity-GRPO: Optimizing Multi-Human Identity-preserving Video Generation via Reinforcement Learning}

\author{%
  Xiangyu Meng$^{1}$\thanks{Equal contribution.} \qquad
  Zixian Zhang$^{2}$\footnotemark[1] \qquad
  Zhenghao Zhang$^{1}$\thanks{Project lead.} \\
  \textbf{Junchao Liao}$^1$ \qquad
  \textbf{Long Qin}$^1$ \qquad
  \textbf{Weizhi Wang}$^1$ \\
  $^1$Alibaba Group \qquad $^2$Fudan University \\
}

\begin{document}

\maketitle

\begin{abstract}

While advanced methods like VACE and Phantom have advanced video generation for specific subjects in diverse scenarios, they struggle with multi-human identity preservation in dynamic interactions, where consistent identities across multiple characters are critical. 
To address this, we propose Identity-GRPO, a human feedback-driven optimization pipeline for refining multi-human identity-preserving video generation. First, we construct a video reward model trained on a large-scale preference dataset containing human-annotated and synthetic distortion data, with pairwise annotations focused on maintaining human consistency throughout the video. We then employ a GRPO variant tailored for multi-human consistency, which greatly enhances both VACE and Phantom. 
Through extensive ablation studies, we evaluate the impact of annotation quality and design choices on policy optimization. Experiments show that Identity-GRPO achieves up to 18.9$\%$ improvement in human consistency metrics over baseline methods, offering actionable insights for aligning reinforcement learning with personalized video generation. Code and weights are publicly available at \href{https://ali-videoai.github.io/identity_page}{Identity-GRPO}. 

% To address this, we propose a human feedback-driven optimization pipeline for refining multi-human identity-preserving video generation. First, we construct a video reward model trained on a large-scale preference dataset containing human-annotated and synthetic distortion data, with pairwise annotations focused on maintaining human consistency throughout the video. We then introduce Identity-GRPO, a GRPO variant tailored for multi-human consistency, which greatly enhances both VACE and Phantom. 
\end{abstract}

\section{Introduction}
The scalability of diffusion transformer architectures~\cite{peebles2023scalablediffusionmodelstransformers} with full 3D attention has propelled the field toward high-quality visual content generation, enabling a broad spectrum of downstream applications, such as identity-preserving video generation or motion-controlled video generation~\cite{zhang2025tora}. Identity-preserving video generation, which aims to create high-fidelity videos with consistent human identity, has become a particularly prominent direction. Early methods like ConsisID~\cite{yuan2025identity} and MovieGen~\cite{polyak2024movie} demonstrated excellence in single-identity video generation. More recently, advancements such as ConceptMaster~\cite{huang2025conceptmaster}, Video Alchemist~\cite{chen2025multi}, Tora2~\cite{zhang2025tora2}, Phantom~\cite{liu2025phantom}, SkyReels-A2~\cite{fei2025skyreels}, and VACE~\cite{jiang2025vace} have extended this paradigm to multi-human generation, fundamentally transforming human-centric content creation pipelines. 

However, for  multi-human identity-preserving video generation (MH-IPV) task, models must simultaneously satisfy complex interactive instructions from text prompts while maintaining identity consistency across the entire video sequence. Even state-of-the-art models like VACE~\cite{jiang2025vace} and Phantom~\cite{liu2025phantom} often erroneously prioritize overall composition similarity over individual identity preservation. For example, these models may swap facial features between characters to fulfill a prompt (e.g., "Two people dancing with synchronized movements but distinct outfits"), resulting in coherent motion patterns but catastrophic identity misalignment. 

Recent advances in Text-to-Video (T2V) generation have demonstrated that diffusion transformer models can better follow complex text prompts through post-training stages involving Reinforcement Learning from Human Feedback (RLHF)~\cite{ouyang2022training}. Methods like DiffusionDPO~\cite{wallace2024diffusion} and DenseDPO~\cite{wu2025densedpo}, adapted from Direct Preference Optimization (DPO)~\cite{rafailov2023direct}, offer a simpler alternative to RLHF by directly optimizing policies that align with human preferences under classification objectives. More recently, FlowGRPO~\cite{liu2025flow} and DanceGRPO~\cite{xue2025dancegrpo} have shown more promising results in aligning video outputs with human preferences through Group Relative Policy Optimization (GRPO)~\cite{guo2025deepseek}.

Despite the success of T2V alignment strategies, applying GRPO to identity-preserving video generation remains underexplored. A critical bottleneck lies in the absence of fine-grained reward models that can disentangle identity preservation from dynamic motion requirements. For T2V, high-level semantic guidance from text allows reward metrics like HPS-v2.1~\cite{wu2023better}, CLIP score~\cite{radford2021learning}, and VideoAlign~\cite{liu2025improving} to effectively capture human preferences. However, in multi-human identity-preserving video generation scenarios, where entities must maintain unique visual characteristics while adhering to complex spatial-temporal interactions, existing reward signals like ArcFace~\cite{deng2019arcface} exhibit high correlations with non-identity-related factors across frames, leading to what we term the "copy-and-paste" effect. Additionally, multimodal conditioning inputs, which consist of multiple reference images paired with corresponding text prompts, introduce significant variance in GRPO training due to substantial differences among group samples. 

In this paper, we propose IdentityGRPO, the first preference-driven alignment strategy for MH-IPV scenarios. To address the critical challenge of reward modeling in this domain, we construct a preference-based multi-human similarity dataset (approximately 15k annotated examples) through a hybrid pipeline that strategically curates and evaluates generated video pairs from five advanced video generation models using a semi-automated framework combined with human labeling. This approach ensures scalability by surpassing manual annotation limits while maintaining strict alignment with human preferences via quality-controlled filtering. Leveraging this dataset, we train a specialized reward model capable of capturing fine-grained identity-consistent quality differences between paired video samples. To optimize practical implementation, we systematically evaluate key hyperparameters, including group size, clip ratio, prompt design, and initialization noise, and identify the most effective configuration for this task. Our experiments demonstrate that this reward model successfully aligns both VACE~\cite{jiang2025vace} and Phantom~\cite{liu2025phantom} models with human preference criteria, establishing a robust foundation for preference-driven MH-IPV.The contributions are summarized as below: 
\begin{itemize}
    \item We construct a large-scale high-quality annotated dataset (15k examples) for multi-human identity-preserving video generation, synthesized from five advanced video generation models. This dataset serves as a foundational resource for evaluating identity-preservation capabilities.
    \item We propose an identity consistency reward model and systematically investigate GRPO training configurations, analyzing the impact of design choices on reward signal effectiveness in multi-human scenarios. 
    \item Our comprehensive experiments demonstrate that IdentityGRPO outperforms baseline methods (VACE and Phantom) by up to 18.9\% and 6.5\% on identity-consistency metrics, providing novel insights into the integration of reinforcement learning with customized visual synthesis for complex multi-human generation tasks.
\end{itemize}

\section{Related Work}

\subsection{Multi-human Identity-preserving Video Generation}
With the advancement of video foundational models~\cite{wan2025wan,kong2024hunyuanvideo}, fine-grained identity-preserving video generation becomes possible. 
% Methods such as~\cite{he2024id,yuan2025identity} integrate identity-specific features extracted from reference images into video generative models via the optimization of auxiliary conditioning modules. But they only customize single identity in generated video. 
% Recently, some approaches including SkyReels-A2~\cite{fei2025skyreels}, HunyuanCustom~\cite{hu2025hunyuancustom}, Phantom~\cite{liu2025phantom} and VACE~\cite{jiang2025vace}, have extended video customization from single identity to multiple identities. 
Recently, SkyReels-A2~\cite{fei2025skyreels} proposed an image-text joint embedding model to inject multi-element representations into existing video foundation models, to effectively learn cross-modal data form integration. Tora2~\cite{zhang2025tora2} introduced a decoupled personalization extractor that generated comprehensive personalization embeddings for multiple open-set entities. HunyuanCustom~\cite{hu2025hunyuancustom} designed a text-identity image fusion module based on LLM for enhanced multi-modal understanding. VACE~\cite{jiang2025vace} utilized a context adapter structure to inject different video generation task inputs into the model, allowing it to handle arbitrary video synthesis tasks, such as reference-to-video generation, multi-human identity-preserving video generation.  Phantom~\cite{liu2025phantom} introduced a dynamic information injection scheme, allowing the insertion of one or more reference human, achieving a unified model architecture for single and multi-human identity-preserving video generation. 
% Despite these promising advances, these methods do not leverage human feedback (e.g., via RLHF~\cite{ouyang2022training}) for model optimization. Consequently, they might fail to align the generated results with human perception, leading to sub-optimal outputs. 
% *Mxy* 最后的结论删掉
% \zhenghao{\begin{CJK*}{UTF8}{gbsn} 又下大结论了，下结论没有why \end{CJK*}}

\subsection{Vision-language Reward Models}
Reward models are crucial in aligning video generation models with human preferences. 
Recently, with the advancement of large vision-language models (VLM)~\cite{bai2025qwen2,wang2025internvl3_5}, many methods use VLM to simulate human perception by performing visual quality score regression or preference learning. VideoScore~\cite{he2024videoscore} trained video quality assessment models on human-annotated video labels.  VisionReward~\cite{xu2024visionreward} performs multidimensional evaluation through 64 fine-grained binary visual QA questions, producing human-aligned visual preference scores. VideoAlign~\cite{liu2025improving}  collected large-scale human-annotated video preference datasets and tuned VLM as a multidimensional reward model to evaluate three critical aspects for text-to-video task: visual quality, motion quality, and text-video alignment. AnimeReward~\cite{zhu2025aligning} constructed a large-scale anime video dataset that incorporated human preferences for both visual appearance and visual consistency. 
Beyond these methods, some works such as LiFT~\cite{wang2024lift}, IPO~\cite{yang2025ipo} and UnifiedReward~\cite{wang2025unified}, following VLM-as-a-judge methods~\cite{lin2024criticbench},  leverage the intrinsic reasoning capabilities of VLMs, where VLMs are tuned to generate cirtic reasons followed by predicted  preference labels. Despite these promising advances, these VLM-as-a-judge methods are not accurately predict human preferences in many cases. 
For MH-IPV task, ArcFace~\cite{deng2019arcface} is usually utilized to evaluate the face consistency. However, as mentioned in~\cite{yuan2025opens2v}, the correlation between FaceSim and human perception is limited. In order to overcome the problems existing in the current methods, we construct the first identity consistency reward model for MH-IPV task. 
% FaceSim exclusively computes similarity scores for facial regions, neglecting the alignment between the generated video and the text prompt. This often results in outputs that are unnatural~\cite{yuan2025opens2v} or inconsistent with the textual  prompts.  
% In order to overcome the problems existing in the current methods, we construct the first identity-consist reward model for MH-IPV task. 
% We evaluate not only the consistency of facial regions, but also the overall visual quality and the fidelity to the text prompt. This encourages the identity-preserving  video generation models~\cite{jiang2025vace,liu2025phantom} to create videos that are identity-consistent, high-quality, and strictly adhere to the given prompts. 
% *Mxy* 第二段要缩减，太长了，然后就是说了太多我们的方法，这里不需要这么多。
% \zhenghao{\begin{CJK*}{UTF8}{gbsn} 前面一堆铺垫，没有任何体现出要做这个reward model的意义  \end{CJK*}}

\subsection{Reinforcement Learning for Image and Video Generation}

% 就说我们是第一个做的，不用说他们不行
% 不要随意抨击别人差

To apply Reinforcement Learning from Human Feedback (RLHF) to image and video generation, early methods either directly fine-tuned models using scalar reward signals \cite{Prabhudesai2023AligningTD, Clark2023DirectlyFD, Xu2023ImageRewardLA, prabhudesai2024video} or employed Reward Weighted Regression (RWR) \cite{peng2019advantage, lee2023aligning, furuta2024improving}. Inspired by Proximal Policy Optimization (PPO) \cite{schulman2017proximal}, policy gradient methods were later integrated into diffusion models and demonstrated effectiveness \cite{black2023training, fan2023reinforcement, gupta2025simple, miao2024training, zhao2025score}. However, PPO-based approaches suffer from high computational costs and sensitivity to hyperparameters. To improve training efficiency, Direct Preference Optimization (DPO)-based methods \cite{rafailov2023direct, wallace2024diffusion, dong2023raft, yang2024using, liang2024step, yuan2024self, liu2025videodpo, zhang2024onlinevpo} directly utilize human preference data as learning signals through a supervised loss objective. Recently, GRPO has shown superior performance in complex reasoning tasks. In the domain of image and video generation, Flow-GRPO \cite{liu2025flow} and Dance-GRPO \cite{xue2025dancegrpo} have successfully introduced GRPO into flow matching models, enabling diverse sampling by reformulating the ODE as an equivalent SDE. To improve the efficiency of the optimization process, Mix-GRPO \cite{li2025mixgrpo} introduces a sliding window mechanism, applying SDE sampling and GRPO-guided optimization only within the window, while using ODE sampling outside of it. However, due to the limitations in reward modeling, no prior work has focused on the MH-IPV task. To our knowledge, Identity-GRPO is the first to adapt reinforcement learning to MH-IPV scenarios.

\section{Method}

\subsection{Preference Dataset Construction}
To facilitate the training of our identity-consistent reward model, we construct a preference dataset for identity preservation by leveraging the OpenHumanVid~\cite{li2025openhumanvid} dataset which is a human-centric video generation dataset. Recognizing that manual annotation of preference is prohibitively costly and labor intensive, thereby limiting the scalability of dataset, we curate an extensive, automatically labeled preference dataset to augment the reward model's training process. 

\subsubsection{Data-filtered Pipeline} 
To construct a high-quality dataset for our task, we designed a multi-stage data filtering and augmentation pipeline. Initially, we address scene complexity and identity ambiguity by programmatically filtering the OpenHumanVid dataset. Qwen3~\cite{yang2025qwen3} is employed to parse captions and limit videos to a maximum of three subjects, followed by using Qwen2.5-VL~\cite{bai2025qwen2} to retain only samples with clear, frontal human faces for maximum facial detail preservation. To prevent background interference in reference images, GroundingDINO~\cite{liu2024grounding} and SAM2~\cite{ravi2024sam} are utilized to precisely segment and extract all human subjects. A key challenge in MH-IPV is the "copy-and-paste" problem~\cite{liu2025phantom,yuan2025opens2v}. To mitigate this, we synthesize varied reference images of each subject from multiple perspectives using Flux.1 Kontext~\cite{labs2025flux}. This forces the model to learn a robust identity representation independent of a single pose. The yield rate of the images edited by Flux.1 Kontext exceeds 80\%, meeting our data requirements. The final pipeline yields a curated dataset of triplets formatted as \texttt{<edited reference image, prompt, original video>}. 
% *MXY*删掉copy-paste的解释
% , where models merely copy the reference image, while disregarding the prompts.

\subsubsection{Automatic Annotation} 
We use our data-filtered pipeline to generate inputs for five personalized video models (VACE-1.3B/14B~\cite{jiang2025vace}, Phantom-1.3B/14B~\cite{liu2025phantom} and MAGREF~\cite{deng2025magref}) to synthesize videos. 
From these, two types of preference pairs are constructed: (1) \texttt{<original video, generated video>}. The generated video can be naturally viewed as a degraded representation of the original video. So, the original video is assigned the default preference. 
(2) \texttt{<generated video 1, generated video 2>}. Both videos are synthesized under identical conditions. The preference label is determined by a majority vote over multiple inferences from VLMs~\cite{bai2025qwen2,hong2024cogvlm2,liu2025nvila}. 
To ensure that preferences reflect identity consistency rather than extraneous factors, such as semantic relevance to the prompt, we filter the dataset by using a SOTA multi-modal vector model GME~\cite{zhang2024gme}. For each pair, we compute the text-video similarity for both videos. Pairs with a significant discrepancy in similarity scores are discarded. This process yields 10,000 automatically annotated preference pairs, which we denote as \texttt{Auto-labeled} data. 
% *MXY*直接说生成天然是真实的退化即可 其他的删掉
% the generated videos often exhibit degradation in identity consistency and visual quality when compared to the original counterparts. We therefore posit that the generated video constitutes one degraded representation of its original counterpart
% \zhenghao{\begin{CJK*}{UTF8}{gbsn} ，没有给出这种自动化build intuitive的解释比如预测可以当成是退化，等 \end{CJK*}}

\subsubsection{Human Annotation} 
Pairwise annotation is a well-established method for capturing relative preferences. It has been demonstrated~\cite{liu2025improving} to achieve higher inter-annotator agreement compared to pointwise methods that rely on the assignment of absolute scores. Therefore, in this work, we employ a pairwise human annotation framework to construct our preference dataset. We provide annotators with guidelines to ensure a clear understanding of the preference criteria. The guidelines instructed annotators to determine the preference labels by evaluating each video based on three key points: facial consistency with the reference image, visual quality over the whole video, and alignment with the text prompt. 
In the annotation interface, annotators were shown a set of reference images, a text prompt, and two generated videos based on these conditions. They were asked to select the video that better preserves the subjects' identities, with the options: \textit{A better / Ties / B better}. A "\textit{Ties}" vote indicates no discernible difference in identity preservation. Reference images were sourced from CelebA-HQ~\cite{karras2017progressive} and our filtered OpenHumanVid data, with prompts also from OpenHumanVid. Each pair was evaluated by three annotators, and the final label was determined by a majority vote. Like automatic annotation, we filtered human-labeled dataset by using the GME~\cite{zhang2024gme} model to ensure data quality. 
This process resulted in 5,000 high-quality, human-annotated preference pairs, denoted as \texttt{Human-labeled} data. 
% *MXY*判断的具体标准是什么，先记下但是不用加

% \zhenghao{\begin{CJK*}{UTF8}{gbsn} 给出pair标注比score好的具体cite，要给出这个相似性判别是基于整个video的整体观感以及对prompt的一致性 \end{CJK*}}

\subsection{Identity-consistent Reward Learning}
\subsubsection{Preference Modeling} 
We adopt Qwen2.5-VL-3B\cite{bai2025qwen2} as the reward model. We employ Bradley-Terry-with-Ties (BTT)~\cite{rao1967ties,liu2024reward} as the object function. BTT explicitly models the tied preferences and conforms to the category of labels in our dataset. 
In the identity-preserving video generation, two videos \(y_A\) and \(y_B\) are generated under identical conditions (i.e., the same reference images \(x\) and prompt \(t\)). The BTT model defines the probabilities of each possible preference as follows: 
\begin{equation}
    P(y^A \succ y^B \mid x,t) = \frac{ e^{r(x, t, y^A)} }{ e^{r(x, t, y^A)} + \theta e^{r(x, t, y^B)} }
\end{equation}

\begin{equation}
P(y^B \succ y^A \mid x, t) = \frac{ e^{r(x, t, y^B)} }{\theta e^{r(x, t, y^A)} +  e^{r(x, t, y^B)} }
\end{equation}

\begin{equation}
P(y^B = y^A \mid x, t) = \frac{(\theta^2 - 1) e^{r(x,t, y^A)} e^{r(x, t, y^B)} } { (e^{r(x, t, y^A)} + \theta e^{r(x,t, y_B)})  (\theta e^{r(x, t, y^A)} +  e^{r(x, t, y^B)}) ) },
\end{equation}              % s_A = \exp(r(x, y_A)), s_B = \exp(r(x, y_B))
where \(r\) is the optimized reward function, and \(\theta \ge 1\) is a parameter that controls the tendency towards ties, with a larger \(\theta\) indicating a higher probability of ties. %When \(\theta = 1\), the BTT model is equivalent to the BT model. 
Following prior work~\cite{liu2024reward}, we set \(\theta = 5\) and optimize the BTT model by minimizing negative log-likelihood loss: 
% \begin{align}
% \mathcal{L}_{RM} = 
% &- \mathbb{E}_{(x, y_A, y_B)\sim\mathcal{D}} \Big[ 
%     \mathbb{I}(y_A \succ y_B) \log P(y_A \succ y_B \mid x) \notag \\
% &\quad + \mathbb{I}(y_B \succ y_A) \log P(y_B \succ y_A \mid x) \notag \\
% &\quad + \mathbb{I}(y_A = y_B) \log P(y_B = y_A \mid x) 
% \Big]
% \end{align}
\begin{equation}
\mathcal{L}_{\text{BTT}} = -\mathbb{E}_{(x, t, y^A, y^B) \sim D} \Big[ \begin{matrix} \sum_{i \in \{y^A \succ y^B, y^B \succ y^A, y^A = y^B\}} \end{matrix} \mathbb{I}(i) \log P(i \mid x, t) \Big],
\end{equation}
where \(\mathbb{I}(i)\) denotes the indicator function. 

\subsubsection{Learning with Auto-labeled Data} 
We propose a two-stage training methodology designed to make full use of the automatically annotated data. Our approach first refines the large, automatically labeled dataset through a consistency-based filtering protocol and then employs a joint training procedure with a dynamic, smooth sampling strategy. 
% \zhenghao{\begin{CJK*}{UTF8}{gbsn} 前面废话太多，直接说为了把自动化标注发挥最大化，我们提出2阶段训练方案，上来一顿大贬低你的自动化方案有noise，直接让人提出问题，那什么要用自动化标注的数据\end{CJK*}}

The data refinement process unfolds as follows. A preliminary reward model, denoted as \texttt{RM\_teacher}, is trained exclusively on the high-quality human-annotated dataset (\texttt{Human-labeled} data). This model serves as a proxy for human preferences. \texttt{RM\_teacher} is used to infer preference scores for all pairs in the larger, automatically labeled dataset (\texttt{Auto-labeled} data). We filter the \texttt{Auto-labeled} dataset by retaining only the samples where the preference predicted by \texttt{RM\_teacher} aligns with the original automatic label. This filtering protocol yields a refined dataset, \texttt{Filtered auto-labeled} dataset, which comprises approximately \text{48\%} of the original data in \texttt{Auto-labeled} dataset. 

In the training stage, we perform joint training on the \texttt{Human-labeled} dataset and the \texttt{Filtered auto-labeled} dataset. To mitigate abrupt distributional shifts when combining these heterogeneous data sources, we introduce a smooth sampling strategy. 
This strategy employs a cosine scheduling mechanism to dynamically adjust the sampling proportion, \(\alpha_t\), for \texttt{Filtered auto-labeled} dataset at each training step \(t\):
\begin{equation}
    \alpha_t = 0.5 \cdot (1 + \cos(\frac{\pi t}{T})),
\end{equation}
where $t \in [0, T]$, \(T\) is the total number of training steps. The value of \(\alpha_t\) monotonically decreases from 1 to 0 as \(t\) progresses from 0 to \(T\). Consequently, the training curriculum commences with a high proportion of data from \texttt{Filtered auto-labeled} dataset and gradually transitions to prioritize the high-fidelity, human-annotated data from \texttt{Human-labeled} dataset. This method ensures training stability and maximizes the utility of both data sources. Our experimental results validate that this approach culminates in a superior-performing identity-consistent reward model.

\subsection{Identity-GRPO Training} 
In this section, we first introduce the preliminary concepts of reinforcement learning, then present the sampling process of Identity-GRPO, and finally describe the method we propose to address the training challenges of GRPO in the MH-IPV task.

\subsubsection{Reinforcement Learning}
Following DDPO \cite{black2023training}, the denoising process of the rectified flow can be formulated as a Markov Decision Process (MDP):
\begin{equation}
\begin{aligned}
& s_t \triangleq (\boldsymbol{c},t,\boldsymbol{z}_t), \ \pi (a_t|s_t)\triangleq p_\theta (\boldsymbol{z}_{t-1}|\boldsymbol{z}_t,\boldsymbol{c}), \ P(s_{t+1}|s_t,a_t) \triangleq (\delta_{\boldsymbol{c}}, \delta_{t-1}, \delta_{\boldsymbol{z}_{t-1}}) 
\\
& a_t \triangleq \boldsymbol{z}_{t-1}, \  R\left({s}_{t}, {a}_{t}\right) \triangleq\left\{\begin{array}{ll}
r\left(\boldsymbol{z}_{0}, \boldsymbol{c}\right), & \text { if } t=0 \\
0, & \text { otherwise }
\end{array}\right., \ \rho_0(s_0)\triangleq (p(\boldsymbol{c}),\delta _T, \mathcal{N}(\mathbf{0}, \mathbf{I})),
\end{aligned}
\end{equation}
where $ s_t $ is the state at step $ t $, $ \pi (a_t|s_t) $ is the policy, $ P(s_{t+1}|s_t,a_t) $ is the deterministic transition, $ a_t $ is the action, $ R\left({s}_{t}, {a}_{t}\right) $ is the reward which is only given at the final step, and $ \rho_0(s_0) $ is the initial state distribution.

In Identity-GRPO, the generative model samples a set of videos $ \{ \boldsymbol{z}_0^i \}_{i=1}^G$ from noise samples $ \{ \boldsymbol{z}_1^i \}_{i=1}^G$. Our reward model then assigns scores $ r(\boldsymbol{z}_0^i,\boldsymbol{c})$ to this set of generated videos, and the advantage of each sample is computed by:
\begin{equation}
\hat{A}_t^i=\frac{r(\boldsymbol{z}_0^i,\boldsymbol{c})-\text{mean}(\left \{ r(\boldsymbol{z}_0^i,\boldsymbol{c}) \right \}_{i=1}^G )}{\text{std}(\left \{ r(\boldsymbol{z}_0^i,\boldsymbol{c}) \right \}_{i=1}^G )}.
\end{equation}
Then the policy model is optimized by maximizing the following objective function:
\begin{equation}
    \mathcal{J}(\theta)=\mathbb{E}_{\left \{ \boldsymbol{z}^i \right \}_{i=1}^G \sim \pi_{\theta_{\text{old}}}(\cdot|\boldsymbol{c}) }\frac{1}{G} \sum_{i=1}^{G} \frac{1}{T} \sum_{t=0}^{T-1}\left(\min \left(\rho_{t}^{i}(\theta) \hat{A}_{t}^{i}, \operatorname{clip}\left(\rho_{t}^{i}(\theta), 1-\varepsilon, 1+\varepsilon\right) \hat{A}_{t}^{i}\right)\right),
\end{equation}
where $ \rho_{t}^{i}(\theta)=\frac{p_{\theta}\left(\boldsymbol{z}_{t-1}^{i} \mid \boldsymbol{z}_{t}^{i}, \boldsymbol{c}\right)}{p_{\theta_{\text {old }}}\left(\boldsymbol{z}_{t-1}^{i} \mid \boldsymbol{z}_{t}^{i}, \boldsymbol{c}\right)} $, and $ \varepsilon $ is a hyper-parameter for clip-range. For the sake of notational simplicity, the KL divergence term is omitted in the formulation.

\subsubsection{Video Sampling}
In the current field of image and video generation, flow matching models have become the dominant approach due to their solid theoretical foundation and exceptional performance. Given a prompt $ \boldsymbol{c}_p $ and its corresponding reference images $ \boldsymbol{c}_r $ as conditions, flow matching models use transformer framework to predict the velocity field $ \boldsymbol{v}_\theta(\boldsymbol{z}_t, \boldsymbol{c},t) $ and:

\begin{equation}
\begin{aligned}
    \boldsymbol{c} &= \{\boldsymbol{c}_p,\boldsymbol{c}_r\} \\
    \text{d}\boldsymbol{z}_t&=\boldsymbol{v}_\theta(\boldsymbol{z}_t, \boldsymbol{c},t)\text{d}t, t\in [0,1].
\end{aligned}
\end{equation}

Following previous works \cite{liu2025flow, xue2025dancegrpo}, we derive the corresponding reverse SDE formulation:
\begin{equation}
    \text{d}\boldsymbol{z}_t= \left[ \boldsymbol{v}_\theta(\boldsymbol{z}_t, \boldsymbol{c},t)+\frac{\sigma^2_t}{2t}(\boldsymbol{z}_t+(1-t)\boldsymbol{v}_\theta(\boldsymbol{z}_t, \boldsymbol{c},t))  \right]\text{d}_t+\sigma_t\text{d}\boldsymbol{w},
\end{equation}
where $ \sigma_t $ introduces the stochasticity during sampling, and we use $ \sigma_t=t $ in our paper.

\subsubsection{Training Stability Strategies}
How to sample a suitable set of videos for subsequent advantage calculation is crucial to the success of GRPO training. However, in MH-IPV, compared to the T2V task, the model input includes multiple modalities, which introduces significant variance and makes it difficult for sampled videos to support stable GRPO training. To address this, we propose two strategies in Identity-GRPO: prompt finetuning and initial noise differentiation. In addition, the variance between different modalities requires a larger number of videos to be used in a single parameter update — analogous to using a larger batch size in standard training. Otherwise, GRPO training is prone to instability or collapse.

\textbf{Prompt Finetuning.}
In MH-IPV, different models exhibit varying sensitivity to the discrepancy between the prompt and the reference image. For example, VACE tends to follow the content of the prompt, whereas Phantom tends to preserve the content of the reference image. This discrepancy leads to reward models being unable to provide reasonable outputs, introducing significant instability during GRPO training. To address this issue, we employed Qwen2.5-VL-7B~\cite{bai2025qwen2} to generate prompts that contain accurate descriptions of the characters in the reference images.

\textbf{Initial Noise Differentiation.}
In MH-IPV, due to the constraints imposed by the reference image, it is difficult to create significant differences in identity consistency among videos within the same group solely relying on the randomness introduced by the SDE. This limits the exploration space required for effective reinforcement learning training. Therefore, during sampling, we employ different initialization noises to amplify the diversity between generated videos.

\textbf{A Larger Video Number.} 
In Identity-GRPO, due to the variance between different modalities, using a limited number of videos in a single parameter update can lead to training instability or collapse. To increase the video number under limited computational resources, we reduce the resolution and number of frames in the sampled videos. Experiments demonstrate that this approach effectively stabilizes the training of Identity-GRPO.

\section{Experiments}

\subsection{Reward Learning}
\textbf{Training Setting.}  
We use LoRA~\cite{hu2022lora} to update the identity-consistent reward model. The vision encoder is also optimized to capture fine-grained details of human identity. Empirically, we set the learning rate to \text{2e-6}. The global batch size is set to 32. We sample videos at 2 fps, with a resolution of approximately 832 \(\times\) 480 pixels, which is the default resolution of VACE~\cite{jiang2025vace} and Phantom~\cite{liu2025phantom}. 

\textbf{Main Results.}  
We establish a preference benchmark for MH-IPV, which comprises 500 video samples meticulously annotated by human evaluators. 
We compare our reward model against ArcFace~\cite{deng2019arcface}, Qwen2.5-VL~\cite{bai2025qwen2} and InternVL3.5~\cite{wang2025internvl3_5}. 
% For VLMs~\cite{bai2025qwen2,wang2025internvl3_5}, we assess their ability to predict preferences for identity-preserving video generation, evaluating the accuracy in both zero-shot and few-shot scenarios. 
The quantitative results are presented in Table~\ref{tab:rm1}. VLMs, such as Qwen2.5VL and InternVL3.5, fail to effectively predict human preferences regarding subject consistency. This is particularly evident in small-scale model, such as the Qwen2.5VL-3B, which serves as our base model. Furthermore, the accuracy of ArcFace~\cite{deng2019arcface} is insufficiently high (below 0.8), indicating its poor alignment with human perception and rendering it unsuitable as a reward signal for human feedback. In contrast, our model outperforms all other methods, showcasing its effectiveness in evaluating identity consistency in MH-IPV task. 
% reward结果中 不说zeroshot和fewshot，用括号表示fewshot增加的量。

\begin{table}[!t]
\centering
\caption{Preference accuracy on our multi-human identity-preserving preference benchmark. \textcolor{red}{Red-colored} font indicates improvement over the baseline when using few-shot examples.}
\begin{tabular}{cccccc}
\toprule
            & ArcFace & Qwen2.5VL-3B & Qwen2.5VL-72B & InternVL3.5-38B & Ours \\ 
\midrule
Accuracy    &  $0.772$   &  $0.430_{\textcolor{red}{+4.4\%}}$    &   $0.657_{\textcolor{red}{+3.6\%}}$  & $0.685_{\textcolor{red}{+2.0\%}}$  & $0.890$ \\
% Few-shot  &   -                 &   0.474  &    0.693  & 0.665  &   -  \\
\bottomrule
\end{tabular}
\label{tab:rm1}
\end{table}

\begin{table}
    \centering
    \caption{Ablation study on data source and data sampling strategies.}
\begin{tabular}{lccccccc}
\toprule
% \midrule
Human-labeled data & \checkmark & & \checkmark & & \checkmark & \checkmark & \checkmark  \\ 
Auto-labeled data & & \checkmark & \checkmark & & & &  \\ 
Filtered auto-labeled data & & & & \checkmark & \checkmark & \checkmark & \checkmark \\ 
\midrule
Random Sampling  & \checkmark & \checkmark & \checkmark & \checkmark & \checkmark & &  \\ 
 % Uniform Sampling & & & & & & \checkmark &  & \\ 
Staged Sampling  & & & & &  & \checkmark & \\  
Smooth Sampling & & & & & &  & \checkmark \\ 
\midrule
Accuracy & \(0.853\) & \(0.664\) & \(0.812\) & \(0.785\) & \(0.877\)   & \(0.824\) & \(0.890\) \\ 
% & 0.867
\bottomrule
\end{tabular}
    \label{tab:rm2}
\end{table}

\textbf{Ablation Study.}  
\label{sec:rm_ab}
As shown in Table~\ref{tab:rm2}, we analyzed the effectiveness of human-annotated data and filtered auto-labeled dataset.
The model trained exclusively on human annotated data achieves an accuracy of 0.853, which we establish as our baseline. In contrast, when trained solely on the raw, unfiltered auto-labeled data, the model's accuracy drops significantly to 0.664. After applying our filtering process, training on the filtered automated data yields an accuracy of 0.785, a substantial improvement of 0.12. Furthermore, when augmenting the human-annotated dataset with automated data, the model's accuracy reaches 0.877 with filtered auto-labeled data, surpassing the baseline. Conversely, using unfiltered automated data results in a lower accuracy of 0.812. These findings collectively underscore the necessity of our data filtering mechanism of auto-labeled data for improving data quality and subsequent model performance. 
% 不要说我们的自动化数据质量差，直接接上后面的句子说明过滤的意义就可以了

During the training phase, we explored several dynamic sampling strategies to effectively combine the filtered auto-labeled dataset and human-annotated dataset. These sampling strategies include: 
(1) \texttt{Random Sampling} denotes that the two datasets are pooled and samples are drawn randomly from the combined collection for training. 
% This strategy is a common way to utilize two datasets. 
% (2) \texttt{Uniform Sampling}, which denotes each training batch is constructed with a 50/50 split, comprising an equal number of samples from Filtered auto-labeled Data and Human-labeled Data. 
(2) \texttt{Staged Sampling} denotes that training stage is divided into two sequential stages of equal length in terms of training steps. The first stage involves training the model exclusively on a large-scale, filtered auto-labeled dataset. The objective of this stage is to enable the model to learn generalizable knowledge and robust feature representations of preferences. In the second stage, the training curriculum shifts to a smaller, high-quality dataset annotated by humans. This stage is designed to align the model's outputs with real human preferences, leveraging the precision of human-provided data. 
(3) \texttt{Smooth Sampling}, which denotes this approach utilizes a probabilistic cosine scheduling mechanism to dynamically adjust the sampling ratio of Filtered auto-labeled and Human-labeled data at each training step. The proportion of samples from Filtered auto-labeled data is high at the beginning of training and gradually shifts towards a higher proportion of samples from Human-labeled data. 
The results in Table~\ref{tab:rm2} indicate that \texttt{Smooth Sampling} outperforms all the other three strategies, with a notable accuracy improvement of 0.066 over \texttt{Staged Sampling}. This is because \texttt{Smooth Sampling} mitigates the abrupt distributional shift that occurs in \texttt{Staged Sampling} strategy. In the latter, this shift causes the parameters from the first stage to become sub-optimal for the new data distribution in the second stage, forcing the model to learn the new feature space and leading to a periodic performance degradation.

\subsection{GRPO Training}

\textbf{Experiment Setup.}
 We utilized the reference images from the preference dataset we constructed, along with the corresponding prompts generated by our prompt finetuning strategy, for GRPO training. The test set consists of 100 samples, which are sufficient to cover a diverse range of scenarios. For quantitative evaluation, we compare the identity consistency reward score (abbreviated as ID-Consistency) produced by the reward model. 
 To evaluate visual quality and text relevance, we use two evaluation metrics from OpenS2V~\cite{yuan2025opens2v}: AestheticScore~\cite{improved-aesthetic-predictor} and GmeScore~\cite{zhang2024gme}. 
 Additionally, we conduct a user study in which participants are asked to select videos with higher ID-Consistency between baseline and Identity-GRPO. The winning rate is then calculated based on user preferences. 
 % We did not use existing face similarity metrics like ArcFace \cite{deng2019arcface} in the MH-IPV task, as they correlate with non-identity-related factors across frames and fail to capture overall video consistency with the reference image and prompt. 

\textbf{Training details.}
Following previous works \cite{liu2022flow, xue2025dancegrpo}, we employ a reduced sampling timestep during training. Specifically, we use 25 steps for sampling and 50 steps for evaluation. The experiments were conducted on 8 A100 GPUs. The sampled videos have a frame length of 33, with a resolution of 416 $ \times $ 240. The number of sampling groups is set to 16, with a group size $G = 8$. The clip range $\varepsilon $ is set to $ 1 \times 10^{-3}$. We incorporate LoRA with $\alpha = 32$ and rank $= 32$.

\begin{table}[t]
    \centering
    \caption{Evaluation results on our test set. \textbf{Bold}: Best Performance.
    }
    \begin{tabular}{lcccc}
        \toprule
        \small{Method} & \small{ID-Consistency \textuparrow} & \small{Aesthetics \textuparrow} & \small{GmeScore \textuparrow} & \small{Winning Rate \textuparrow} \\
        \midrule
        \small{VACE-1.3B} & \small{2.606} & \small{45.58$\%$} & \small{67.98$\%$} & \small{24$\%$} \\
        \small{VACE-1.3B+Identity-GRPO} & \small{\textbf{3.099}} & \small{\textbf{47.56$\%$}} &  \small{\textbf{68.35$\%$}} & \small{\textbf{76$\%$}} \\
        \midrule
        \small{Phantom-1.3B} & \small{3.809} & \small{44.13$\%$} & \small{67.56$\%$} & \small{37$\%$} \\
        \small{Phantom-1.3B+Identity-GRPO} & \small{\textbf{4.056}} & \small{\textbf{47.03$\%$}} & \small{\textbf{68.47$\%$}} & \small{\textbf{63$\%$}} \\
        \bottomrule
    \end{tabular}
    \label{grpo-tab-main}
\end{table}

\begin{figure}[ht]
    \centering
    \includegraphics[width=\linewidth]{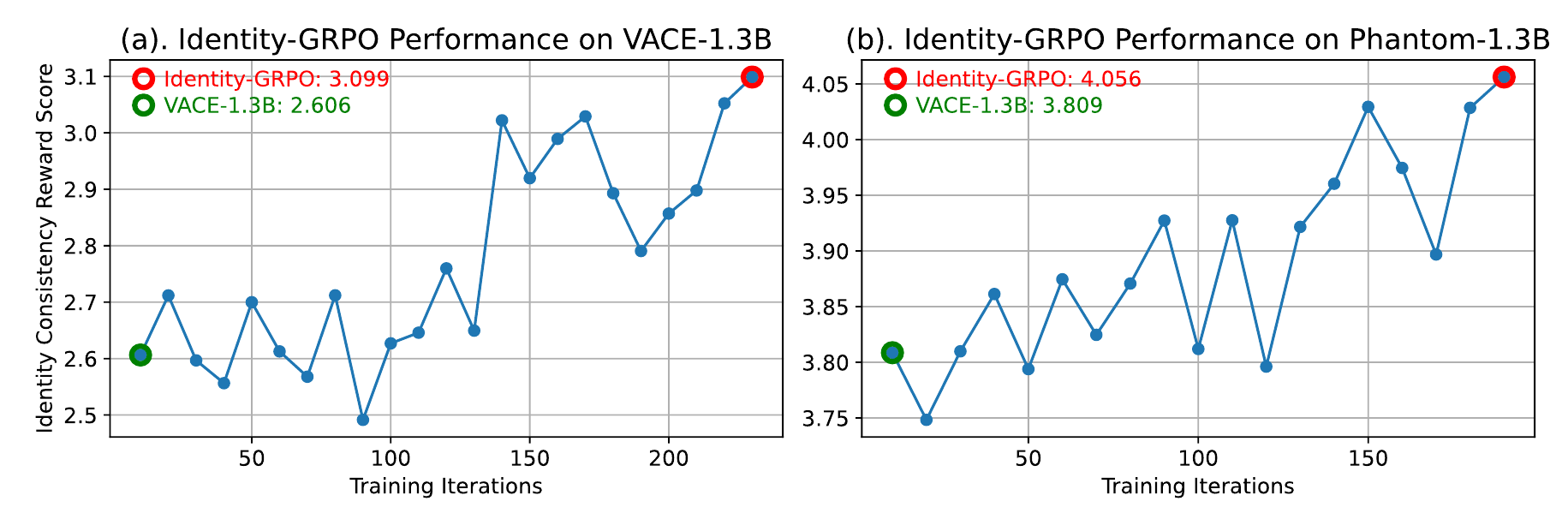}
    \caption{(a) and (b) respectively show the performance curves of Identity-GRPO on VACE-1.3B and Phantom-1.3B. Both exhibit a clear upward trend.}
    \label{curve}
\end{figure}

\textbf{Main Results.}
Figure \ref{curve} illustrates the performance of Identity-GRPO on the test set throughout the training process. Both models exhibited a clear upward trend during the GRPO training process. As detailed in Table \ref{grpo-tab-main}, Identity-GRPO improved ID-Consistency of VACE-1.3B and Phantom-1.3B by up to 18.9\% and 6.5\%, respectively. The advantage of Identity-GRPO was also evident in the user study, achieving winning rates of 76\% and 63\%. 
Furthermore, Identity-GRPO enhances the aesthetic scores of both VACE-1.3B and Phantom-1.3B, while the GME scores remain largely stable throughout the fine-tuning process. These results demonstrate that Identity-GRPO effectively strengthens identity preservation without compromising visual quality or prompt-following performance.

\begin{figure}
    \centering
    \includegraphics[width=\linewidth]{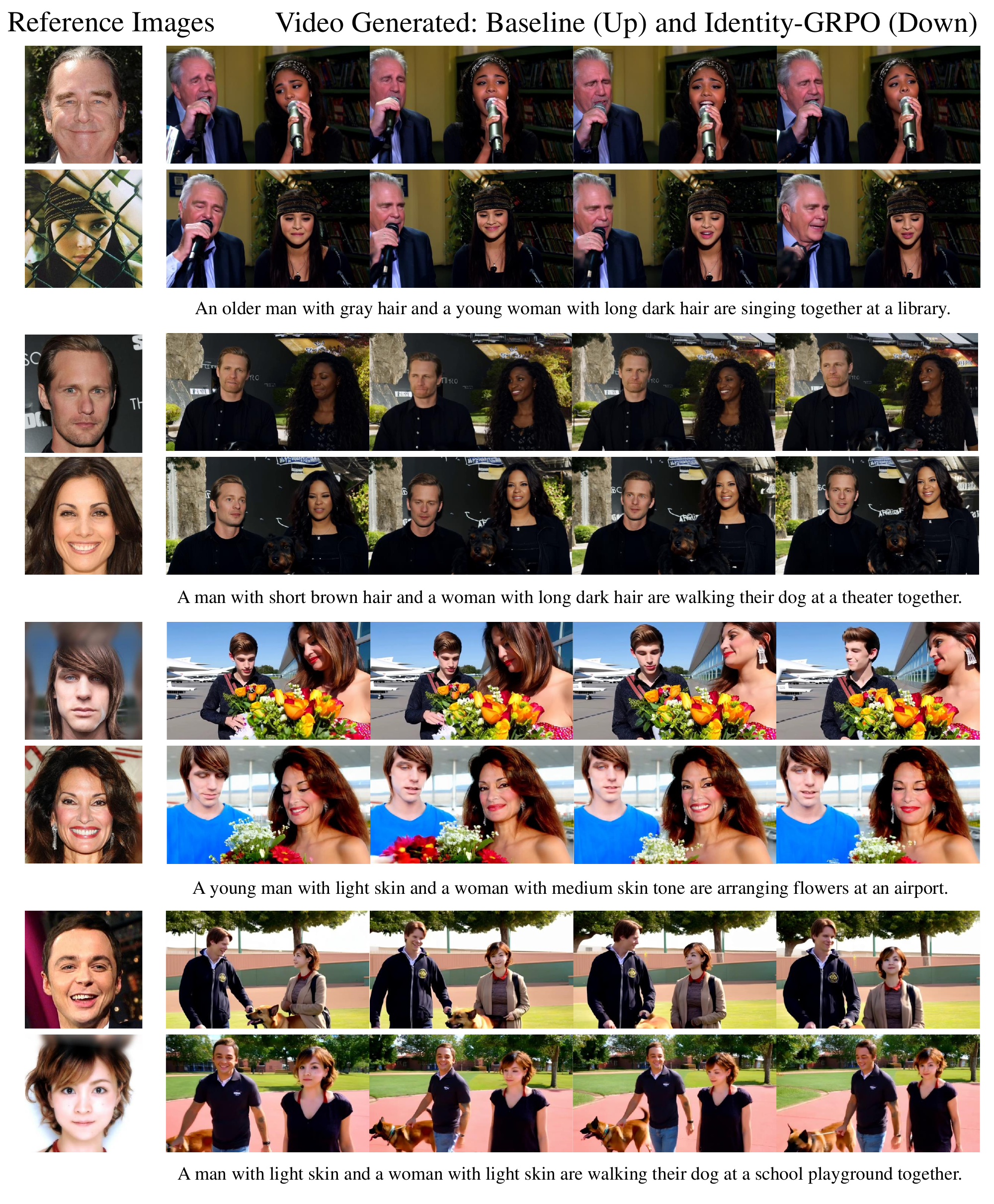}
    \caption{Visualization results for qualitative analysis. The first two groups show a comparison between VACE-1.3B and VACE-1.3B+Identity-GRPO, while the last two groups compare Phantom-1.3B with Phantom-1.3B+Identity-GRPO. 
    In each group, the first row presents the results from the baseline model, and the second row shows the results generated by Identity-GRPO.
    }
    \label{grpo-vis-fig}
\end{figure}

Qualitative comparison results are provided in Figure \ref{grpo-vis-fig}. In several cases, the baseline model produced outputs that clearly mismatched the reference image, whereas Identity-GRPO consistently maintained high identity alignment.

\textbf{Ablation Study.}
As discussed in the Method section, due to the multimodal input conditions in the MH-IPV task, using a large number of videos in each parameter update, along with diverse initial noises, is crucial for achieving stable GRPO training. Therefore, we conducted ablation studies on the impact of the video number and the initial noise on the VACE-1.3B training process of Identity-GRPO. The number of videos is equal to the number of sampling groups multiplied by the group size. In our experiments, we fix the group size at 8 and vary the number of sampling groups. As shown in Table~\ref{grpo-tab-ablation}, which presents the final results under four training settings. When the video number is insufficient, GRPO training becomes unstable and consistently fluctuates. Using the same initialization noise when sampling a group of videos restricts the exploration space of GRPO, preventing the reward from increasing.

Leveraging our curated high-quality dataset, we conducted a comparative analysis of two post-training methodologies: Supervised Fine-Tuning (SFT) and Identity-GRPO. To evaluate identity preservation, we employed the ArcFace similarity score as metrics. As shown in Table~\ref{grpo-tab-ablation-sft}, the baseline VACE-1.3B achieved a score of 0.235, and SFT reached 0.261, whereas our proposed Identity-GRPO yielded a superior score of 0.298. These results demonstrate that Identity-GRPO significantly outperforms SFT in maintaining identity consistency, validating its efficacy in enhancing the generation quality of video models when trained on high-quality datasets.

\begin{table}[t]
    \centering
    \caption{Ablation study on group number and initial noise. \textbf{Bold}: Best Performance.}
    \begin{tabular}{c|cc}
        \toprule
        Group Num & Different Initial Noise & ID-Consistency\\
        \midrule
        4 &  & 2.588 \\
        4 & \checkmark & 2.749 \\
        16 &  & 2.718 \\
        16 & \checkmark & \textbf{3.099} \\
        \bottomrule
    \end{tabular}
    \label{grpo-tab-ablation}
\end{table}

\begin{table}[t]
    \centering
    \caption{Ablation study on different post-training methodologies. \textbf{Bold}: Best Performance.}
    \begin{tabular}{cccc}
        \toprule
         & VACE-1.3B & Supervised Fine-Tuning & Identity-GRPO \\
        \midrule
        ArcFace similarity & 0.235 & 0.261 & \textbf{0.298} \\
        ID-Consistency & 2.606 & 2.774 & \textbf{3.099} \\
        \bottomrule
    \end{tabular}
    \label{grpo-tab-ablation-sft}
\end{table}

\section{Conclusion}
This paper presents Identity-GRPO, the first preference-driven alignment strategy for multi-human identity-preserving video generation (MH-IPV), addressing critical challenges in maintaining identity consistency during complex spatial-temporal interactions. By constructing a large-scale annotated dataset  through a hybrid semi-automated labeling pipeline, we enable the training of a fine-grained identity-consistency reward model tailored to disentangle identity preservation from dynamic motion requirements. Our systematic evaluation of GRPO training configurations identifies optimal hyperparameters for MH-IPV scenarios, demonstrating that Identity-GRPO outperforms state-of-the-art baselines (VACE and Phantom) by 18.9$\%$ and 6.5$\%$ in identity-consistency metrics. These contributions establish a robust foundation for aligning reinforcement learning with human-centric video generation, offering actionable insights for advancing scalable, identity-preserving multi-human content creation. 

\bibliographystyle{plainnat}    % xxx
\bibliography{neurips_2025}     % xxx

\end{document}